\documentclass{article}

\usepackage[preprint]{neurips_2024}

\usepackage[utf8]{inputenc} 
\usepackage[T1]{fontenc}    
\usepackage{hyperref}       
\usepackage{url}            
\usepackage{booktabs}       
\usepackage{amsfonts}       
\usepackage{nicefrac}       
\usepackage{microtype}      
\usepackage{xcolor}         
\usepackage{amsmath}
\usepackage{amssymb}
\usepackage{nomencl}
\usepackage{glossaries}
\usepackage{xspace}
\usepackage{multirow}
\usepackage[pdftex]{graphicx}
\usepackage{bbding}
\usepackage{pifont}
\usepackage{algorithm}
\usepackage{subfigure}
\usepackage{enumitem}
\usepackage{multicol}
\usepackage{caption}
\usepackage{array}
\usepackage{fp}
\usepackage{makecell}
\usepackage{flushend}
\usepackage{cases}
\usepackage{color}
\usepackage{algpseudocode}
\usepackage{listings}
\usepackage{mathrsfs}
\usepackage{longtable}
\usepackage{colortbl}
\usepackage{bm}

\definecolor{codegreen}{rgb}{0,0.3,0.6}
\definecolor{codegray}{rgb}{0.5,0.5,0.5}

\newcommand{\ie}{\emph{i.e.,}\xspace}
\newcommand{\eg}{\emph{e.g.,}\xspace}
\newcommand{\aka}{\emph{a.k.a.,}\xspace}

\newcommand{\ignore}[1]{}

\usepackage{tcolorbox}
\definecolor{darkorange}{RGB}{255, 140, 0}
\definecolor{lightgreen}{RGB}{145, 204, 117}
\definecolor{lightyellow}{RGB}{250, 200, 88}
\definecolor{lightred}{RGB}{238, 102, 102}
\definecolor{lightblue}{RGB}{115, 192, 222}
\newtcolorbox{promptbox}[2][Prompt]{
colback=black!5!white,
arc=5pt, 
boxrule=0.5pt,
fonttitle=\bfseries,
title=#1, 
before upper={\scriptsize}, fontupper=\fontfamily{ptm}\selectfont,
colframe=#2, 
}

\title{Enhancing LLM Reasoning with Reward-guided \\Tree Search}

\author{%
    Jinhao Jiang$^{1}$\thanks{Equal contribution.}~,
  Zhipeng Chen$^{1*}$,
  Yingqian Min$^{1*}$,\\
  \textbf{Jie Chen$^{1}$,
  Xiaoxue Cheng$^{1}$, 
  Jiapeng Wang$^{1}$,
 Yiru Tang$^{1}$, 
  Haoxiang Sun$^{2}$, 
  Jia Deng$^{1}$,}\\
  \textbf{Wayne Xin Zhao$^{1}$\thanks{Correspondence to Xin Zhao.}~, 
  Zheng Liu$^3$, Dong Yan$^4$, Jian Xie$^4$,} \\
  \textbf{Zhongyuan Wang}$^3$, \textbf{Ji-Rong Wen}$^{1}$
  \\
  $^1$Gaoling School of Artificial Intelligence, Renmin University of China.\\
  $^2$School of Information, Renmin University of China.\\
  $^3$BAAI.\\
   $^4$Baichuan AI.\\
  \texttt{\{jiangjinhao,zhipeng\_chen,yingqianm,jrwen\}@ruc.edu.cn}\\
  \texttt{batmanfly@gmail.com}
}

\begin{document}
Technical Report on Slow Thinking with LLMs: I

\maketitle

\begin{abstract}
Recently, test-time scaling has garnered significant attention from the research community, largely due to the substantial advancements of the o1 model released by OpenAI. By allocating more computational resources during the inference phase, large language models~(LLMs) can extensively explore the solution space by generating more thought tokens or diverse solutions, thereby producing more accurate responses. 
However,  develop an o1-like reasoning approach is challenging, and  researchers have been making various attempts to advance this open area of research.
In this paper, we present a preliminary exploration into enhancing the reasoning abilities of  LLMs through reward-guided tree search algorithms. 
This framework is implemented by integrating the policy model, reward model, and search algorithm. It is primarily constructed around a tree search algorithm, where the policy model navigates a dynamically expanding tree guided by a specially trained reward model. The implemented framework is denoted as \textbf{STILL-1} (\underline{S}low \underline{T}h\underline{i}nking with \underline{LL}Ms), marking the first model developed by our project, ``\emph{Slow Thinking with LLMs}''. 
We thoroughly explore various design considerations necessary for implementing this framework and provide a detailed report of the technical aspects. 
To assess the effectiveness of our approach, we focus on mathematical reasoning tasks and conduct extensive evaluations on four challenging datasets, significantly enhancing the reasoning abilities of LLMs. 
\end{abstract}

\section{Introduction}
In recent years, researchers have made substantial advancements in the capabilities of large language models (LLMs) by scaling both the training data and model parameters in accordance with the (training-time) scaling law~\cite{kaplan2020scaling, henighan2020scaling}.
While LLMs are proficient in processing a wide range of human instructions, their performance remains limited in complex reasoning tasks, such as those encountered in STEM disciplines (including Olympiad-level mathematics, physics, and biology), coding, and medical diagnosis~\cite{glazer2024frontiermath,Jimenez2024swebench,Huang2024OlympicArena}. In response to these limitations, researchers have explored various strategies to improve the complex reasoning abilities of LLMs~\cite{Lewkowycz2022minerva, jiuzhang3.0, wei2023chain,wang2023selfconsistency,Yao2023tree}, including optimizations during both the training and test phases, or a combination of the two. 

Specifically, training-time optimizations typically involve the design of targeted training strategies, utilizing curated datasets or tailored optimization objectives to enhance the capabilities of LLMs (\aka policy models). These strategies range from single-turn methods (\eg direct preference optimization~\cite{rafailov2024direct}) to iterative approaches (\eg self-improvement training~\cite{Yuan2023rft,Zelikman2022star,Hosseini2024vstar}). 
In contrast, test-time scaling expands the reasoning space by generating additional tokens (\eg chain-of-thought reasoning~\cite{wei2023chain}, CoT) or solutions (\eg self-consistency~\cite{wang2023selfconsistency}, SC). This approach aims to enhance problem-solving performance by allowing more inference costs, simulating the slow thinking process of System 2~\cite{daniel2017thinking}. It often integrates search based techniques such as beam search~\cite{Kang2024mindstar} and Monte Carlo Tree Search (MCTS)~\cite{chen2024alphamath,Zhang2024restmcts} to more effectively identify improved solutions. To well support test-time scaling, a capable reward model is often required—either process-based or outcome-based—to guide the learning of the policy model through feedback signals, which is considered essential for search-enhanced reasoning.
In the early stages, training-time optimizations were the primary focus of the research community; however, more recently, test-time scaling has gained greater prominence, particularly following the release of OpenAI's o1 model~\footnote{https://openai.com/o1/}.
o1 has demonstrated notable improvements across a wide range of challenging evaluation benchmarks, spanning from programming to  scientific disciplines. 
According to the official blog, o1 is characterized by conducting long internal thoughts of chains prior to responding,
which allows for performance gains through increased train-time compute (\ie reinforcement learning) and extended test-time compute (\ie thinking during inference).

Despite significant advancements, OpenAI has not disclosed the core technology underlying the o1 model. Consequently, recent research efforts have focused on uncovering the ``secret'' behind its approach to enhancing complex reasoning capabilities of LLMs~\cite{chen2024alphamath,Putta2024agentq,Zelikman2024quiet,Luo2024improve,Hosseini2024vstar}. Generally speaking, most of  these efforts aim to emulate the working mechanisms of AlphaGo (or its enhanced version, AlphaZero) and conceptualize a search-based reasoning framework comprising three key components: the policy model, reward model, and search algorithm. Within this framework, the policy model learns to solve complex tasks by generating reasoning steps (or thoughts) guided by the search algorithm, with its performance further refined and directed by the reward model. These efforts have yielded promising results, though they have not yet reached the level of performance demonstrated by o1.

Implementing a search-based reasoning framework, as demonstrated in prior studies, often entails a multitude of design considerations or technical details, As a result, reproducing an o1-like system is not trivial, even when performance is not the primary objective. A comprehensive report detailing the exploration of various implementation aspects would be useful in guiding future research in this area. In fact, several studies and projects~\cite{zhang2024llamaberry, qin2024o1replicationjourneystrategic} have been dedicated to serving this purpose. 
In parallel with these works, we present a technical report documenting our team's preliminary exploration of a reward-guided tree search framework for improving LLM reasoning. Our framework is primarily constructed around a tree search algorithm, where the policy model navigates a dynamically expanding tree, guided by a specially trained reward model. We denote the implemented framework by \textbf{STILL-1} (\underline{S}low \underline{T}h\underline{i}nking \underline{LL}Ms), which is the first model developed by our project, ``\emph{Slow Thinking with LLMs}''. 
We thoroughly explore the various design considerations necessary to implement this framework and report the technical details.  
To summarize, we list the major attempts at each aspect below:

\begin{itemize}
    \item For the policy model, we investigate how to adapt it to the reasoning format defined by the search tree structure. Additionally, we outline how to perform preference optimization using specially constructed training data, under the guidance of the reward model.

    \item For the reward model, we examine several key design considerations, including the selection of discriminative or generative forms, outcome- or process-supervised training, and ranking- or score-based optimizations. We provide comprehensive training details and also explore how to conduct iterative mutual refinement with the policy model.

    \item For  tree search, we implement MCTS-like algorithm to support the reasoning of the policy model, guided by the reward model. We explore enhancements in both effectiveness and efficiency to make it well-suited for mathematical reasoning tasks.
\end{itemize}

To validate the effectiveness of our implemented framework, we conduct extensive evaluations on four challenging mathematical benchmark datasets: MATH-OAI~\cite{LightmanKBEBLLS24lets}, GSM-Hard~\cite{GaoMZ00YCN23pal}, Olympiad Bench~\cite{HeLBHTSHHHZLQL024olympiad}, and College Math~\cite{TangZWW24mathscale}. Experimental results demonstrate that our reasoning framework significantly enhances the performance of the policy model on these datasets.
Additionally, we conduct an in-depth empirical analysis of the design of the policy model, reward model, and tree search algorithm, aiming to provide researchers with meaningful guidance.

\section{Method}
In this work, we focus on the mathematical domain, specifically addressing mathematical problems described in text. We implement a reward-guided tree search framework, \textbf{STILL-1},  designed to enhance the reasoning capabilities of LLMs. This framework consists of three main components: the policy model, the reward model, and the search algorithm. Within our framework, the policy model generates new reasoning steps based on a partial solution prefix along a given path in the search tree. The search algorithm constructs the search tree to structure the reasoning process, while the reward model provides feedback signals to guide the policy model's actions and the search process. This approach allows the policy model to explore a broader reasoning space in a more informed manner, increasing the likelihood of finding the correct answer. Overall, it trades test time for improved accuracy. In the following sections, we will provide detailed descriptions of the implementation of these three key components.

\label{sec_policy}
\begin{figure}[htbp]
    \centering
    \includegraphics[width=0.95\textwidth]{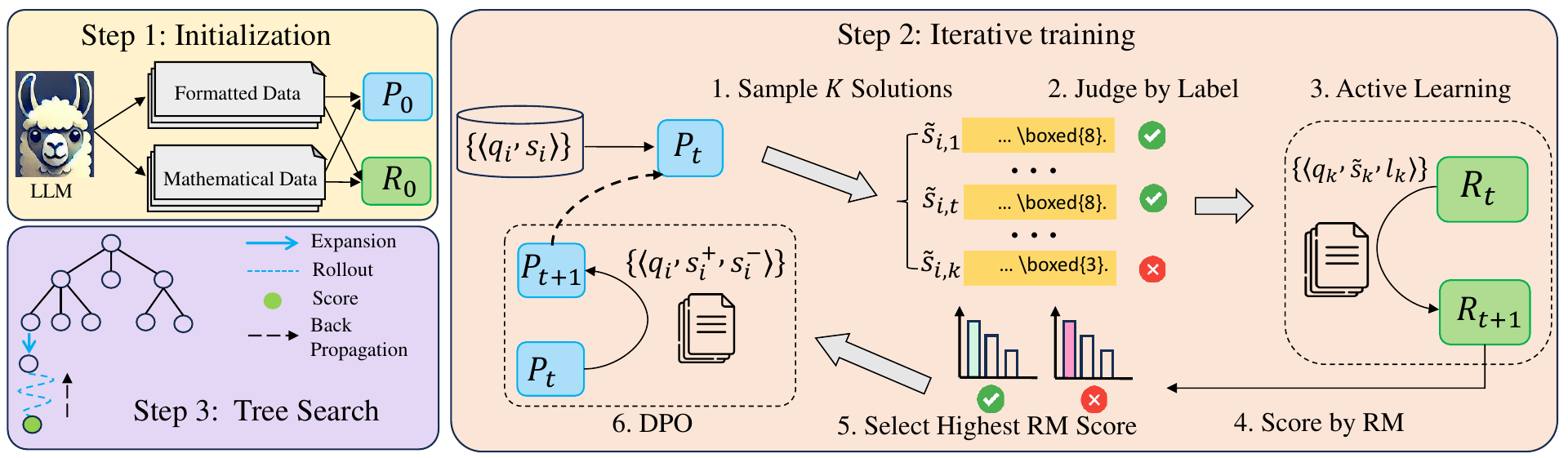} 
    \caption{An overview of our reasoning framework STILL-1. }
    \label{fig:o1}
\end{figure}

\subsection{Policy Model}\label{sec-policy}

In this section, we provide a detailed description of the training process for the policy model, which primarily consists of two steps: instruction tuning for reasoning format adaptation (Section~\ref{sec:format_adaptation}), and preference optimization for policy improvement (Section~\ref{sec:preference_optimization}).

\subsubsection{Instruction Tuning for Reasoning Format Adaptation}
\label{sec:format_adaptation}
As we perform the reasoning using a tree-structured search algorithm, this necessitates adapting the reasoning format of the policy model.

\paragraph{Defining the Reasoning Format.} To define the reasoning format, we first set the granularity of each reasoning step, \ie how each node is represented, framing the multi-step reasoning process as a node-based search within a tree structure.
Previous studies have extensively explored reasoning steps at both the token and sentence levels~\cite{lightman2024step-verify, rafailov2024Qfunc}.
However, in mathematical contexts, a logically complete step in problem-solving may involve multiple sentences. Therefore, we model a complete logical step within the mathematical problem-solving process  as a node in the tree. 
Below is the output template for our reasoning process.

\begin{promptbox}[Step-by-Step Output Format in Our Reasoning Process]{codegray}
\texttt{**Problem Formulation**\\Rephrasing the problem in your own words to capture its essential meaning\\\\ **Step 1: Brief title that summarizes the overall goal of that step**\\Detailed calculation or problem-solving process of this step.\\\\...\\\\ **Step i: Brief title that summarizes the overall goal of that step**\\Detailed calculation or problem-solving process of this step.\\\\...\\\\ **Final Answer**\\Output the final answer in boxed\{\}.}
\end{promptbox}

\paragraph{Supervised Format Following.} As shown in the prompt template, our reasoning format begins by suggesting that the LLM thoroughly understands the problem, followed by a sequence of individual steps, each with a summarization label and a detailed description or calculation. To align with this desired reasoning format, we synthesize formatted data through in-context learning with a more capable LLM, followed by instruction tuning on the policy model. Specifically, using the NuminaMath dataset~\cite{li2024numinamath}, we employ the Qwen2.5-Math-72B-Instruct model~\cite{Yang2024qwen2.5math} to generate formatted solutions for each problem via one-shot in-context learning. After filtering out incorrect solutions, we conduct supervised fine-tuning on the policy model with these instruction data to effectively improve the reasoning abilities while ensuring it adapts well to the reasoning format. We utilize the following prompt template for our policy model during the supervised fine-tuning and later inference.
\begin{promptbox}[Prompt Template for Policy Model]{codegray}
\texttt{Analyze and respond to the following question step by step. Begin by rephrasing the problem in your own words to capture its essential meaning accurately. Then, proceed to solve the problem systematically, ensuring that each step is introduced with a concise heading that summarizes its objective. Follow this with detailed explanations of the calculations or methodologies employed in the problem-solving process. Finally, present the final answer within boxed\{\}.\\ \{Problem\}}
\end{promptbox}

\subsubsection{Preference Optimization for Policy Improvement}
\label{sec:preference_optimization}
By aligning the policy model with the desired reasoning format, we can better control the step-by-step generation of the reasoning process. This alignment facilitates both the execution of tree search algorithms and the assessment by the reward model. Next, we conduct preference optimization on the policy model using feedback from the reward model (Section~\ref{sec-reward}), thereby achieving policy improvement. This process involves two key steps: constructing the training data and performing preference optimization. Next, we introduce the implementation of these two steps.

\paragraph{Training Data Construction.} 
To improve the policy model, we construct the corresponding training data through the following three steps: first, sample multiple solutions for each question from the policy model; second, filter out low-quality solutions; and third, create positive and negative sample pairs based on scores from the reward model and the annotated labels. 
Specifically, given a set of mathematical problems $\mathcal{Q} = \{\langle q_i,s_i\rangle\}_{i=1}^{N}$, we begin by solving each problem $q_i$ using the aforementioned policy model after adapting its reasoning format. During the generation process, step-level diversity is crucial, as sufficiently varied paths are essential for exploring different actions within the tree.
To address this, we increase the temperature parameter to enhance sampling diversity, thereby obtaining more diverse solution paths $\{\tilde{s}_{i,j}\}_{j=0}^{k}$. 
Next, we apply a rule-based approach to remove samples that contain garbled content or deviate from the specified reasoning format. We then annotate the remaining samples with correctness labels by comparing the predicted answers to the ground-truth $s_{i}$, and only retain those problems with both correct and incorrect samples. 
Subsequently, we score each retained sample using the reward model and select the top-ranked correct and incorrect samples to form positive-negative sample pairs. This approach aims to select highly confident positive samples and incorrect samples that are nearly correct, providing more informative signals for preference optimization. 
Similar strategies have been also used in previous work~\cite{Zelikman2022star,Hosseini2024vstar,Yuan2023rft,Zhang2024restmcts}.
Additionally, we sample only one pair of positive and negative examples per problem to increase the diversity of the training data and prevent overfitting. Ultimately, we obtain the training data set $\{\langle q_i, s_i^+, s_i^- \rangle\}_{i=0}^{N}$.

\paragraph{Preference Optimization.}
After obtaining the training data, {we apply the chat template and prompt to the training data} and then conduct preference optimization on the policy model $\mathcal{P}$ using the direct preference optimization algorithm~\cite{rafailov2024direct} as follows:
\begin{align}\small
    \mathcal{L} = - \log \sigma \bigg(\beta \log \frac{\pi(s_i^+|q_i;\Theta)}{\pi(s_i^+|q_i;\Theta_\text{ref})} - \beta \log \frac{\pi(s_i^-|q_i;\Theta)}{\pi(s_i^-|q_i;\Theta_\text{ref})}\bigg),
\end{align}
where $\sigma(\cdot)$ denotes the sigmoid function, $\Theta_\text{ref}$ denotes the parameters of the reference model (\ie the policy model itself), and $\pi$ denotes the generative probability of a sample text given the question and the corresponding model. 

Note that although we describe a single-turn policy improvement process above, it can be easily extended to multi-turn optimization using mutually optimized reward models, which will be detailed in Section~\ref{sec:iterative_training}. 

\subsection{Reward Model}
\label{sec-reward}
In this section, we first identify the key design considerations in implementing the reward model, and then introduce its detailed training process.

\subsubsection{Key Design Considerations in Reward Modeling}
\label{sec-rm_key_issue}

Since the reward model (RM) provides feedback signals to guide the policy model’s reasoning within the search framework, it plays a crucial role in directing the reasoning process of LLMs.  Generally, a reward model can be implemented using various types of machine learning models~\cite{AlphaGo,ppo,Ouyang2022instruct}; however, in our work, we focus on using LLMs as the backbone model. To design an effective RM, we concentrate on the following three key considerations.

\paragraph{Discriminative RM \emph{v.s.} Generative RM.}
When implementing reward models for LLM reasoning, we can consider using either discriminative RM or generative RM.
In existing work~\cite{Wang2024mathshepherd,Ouyang2022instruct,Yang2024qwen2.5math}, discriminative RM has been widely utilized to provide supervision signals, 
which projects the model's hidden state into a score that serves as the supervision signal. In contrast, generative RM takes a specific prompt as input (\ie guidance for evaluation and the text to be evaluated) and generates a descriptive evaluation of the quality for the provided solution, potentially accompanied by associated reward scores~\cite{mahan2024generativerewardmodels,Zhang2024Generative,gpt4}.
A direct implementation of generative RM involves constructing assessment prompts, such as “\texttt{Is the answer correct (Yes/No)?}”, and then using the prediction probabilities of the assessment tokens (\eg the probability of “\texttt{Yes}”) as the reward score. 
Compared to discriminative RM, a potential advantage of generative RM is that it can leverage the learned knowledge of the base model more effectively, as its training process closely aligns with both the pre-training and post-training phases of the model.

\paragraph{Outcome-supervised RM \emph{v.s.} Process-supervised RM.}
The second issue we consider is the granularity of the supervision signals provided by the reward models. Typically, we can use either outcome-supervised reward models~(ORM), which assess the correctness of the entire solution, or process-supervised reward models~(PRM), which evaluate the correctness of intermediate steps in the reasoning process~\cite{LightmanKBEBLLS24lets, Wang2024mathshepherd, Uesato2022solving}. 
To develop these two types of reward models, it requires labeled data at different granularities: solution-level labels and step-level labels. In practice, solution-level labels are relatively easy to obtain when ground-truth answers are available for the tasks, whereas step-level labels are very difficult to annotate. 
To automatically obtain process-supervised signals, we can employ rollout-based methods (\eg Math-Shepherd~\cite{Wang2024mathshepherd}) to derive step-level labels from ground-truth answers. The core idea is to annotate each reasoning step based on whether correct answers can be derived from it through a rollout process. 
However, rollout-based methods typically require more computation time, particularly when the process is repeated multiple times. As a result, this work focuses on training outcome-supervised reward models, primarily relying on outcome-level supervision signals to guide the reasoning process. Interestingly, we find that the reward model trained with solution-level labels also has potential in assessing step-level correctness.

\paragraph{Ranking-based RM \emph{v.s.} Scoring-based RM.}
To train reward models, we can explore different optimization objectives, which are generally either ranking-based or scoring-based. Specifically, a ranking-based RM is trained to identify which candidate response is the best, focusing primarily on the  preference order among candidate solutions. In contrast, a scoring-based RM assigns an absolute reward score for a single response based on its assessed quality level. 
The following equations illustrate the comparison between ranking-based and scoring-based approaches:
\begin{eqnarray}
\text{Rank}(r_1, r_2, r_3) &\rightarrow& r_1 > r_3 > r_2; \\
\text{Score}(r) &\rightarrow& 0.8.
\end{eqnarray}
Since our reasoning framework relies on the guidance of concrete scores for solution evaluation, we adopt the scoring-based optmization objective for training our RM.

To summarize, the key configurations for our reward model are as follows\footnote{Note that this configuration is based on both our empirical experiments and the availability of  computational resources, which may not necessarily be the optimal configuration.}: \emph{generative}, \emph{outcome-supervised}, and \emph{scoring-based}. In Section~\ref{exp-rm}, we will present detailed experimental results comparing the performance of these different configurations. 
The following sections will introduce the details of the training process of reward model.

\subsubsection{Training Data Construction}
\label{sec:data_construction}
To effectively train the reward model, we construct a high-quality dataset with {outcome-level supervision signals}, using rule-based methods to clean and filter the training data.

\paragraph{Data Collection.}
In order to provide accurate guidance to the policy model, we curate the training instances  directly from the content generated by the corresponding policy model.
{
We begin by sampling multiple candidate solutions $\{\tilde{s}_{i,1}, \dots, \tilde{s}_{i,k}\}$ for each problem in the training set. 
}
To ensure diversity among the solutions, we adjust hyperparameters such as \texttt{temperature} and \texttt{top\_p} to promote more varied outputs.
After obtaining the generated solutions for each problem, these candidates are labeled according to the ground-truth answer $s_i$. 
The label $l_{i,j}$ indicates whether the candidate solution $\tilde{s}_{i,j}$ is correct or incorrect. The resulting labeled instances constitute the original training dataset for the reward model, denoted as $\mathcal{D}_{O} = \{ \langle q_i, \tilde{s}_{i,j}, l_{i,j} \rangle \}$. 

\paragraph{Data Cleaning.} Although we adjust the hyperparameters for generation to maintain diversity among the generated solutions, some candidate solutions may still be highly similar, potentially leading to overfitting in the reward model. To address this issue, we perform data cleaning by removing redundant solutions. Specifically, we iteratively eliminate responses that have a high $n$-gram overlap with previously selected responses.
Additionally, we balance the dataset by ensuring an equal number of correct and incorrect solutions for each problem. This is achieved through random sampling, which helps prevent potential bias in the training dataset. After applying the filtering and balancing process, the remaining solutions are used to construct the training instances. The final cleaned training dataset is denoted as $\mathcal{D}_{T} = \{ \langle q_k, \tilde{s}_k, l_k \rangle \}$. 

\subsubsection{Reward Model Training}
\label{reward-model-training-approach}
After constructing the training data, we leverage these data to train the generative reward model $\mathcal{R}$, which can provide outcome-level absolute scores based on the problems and generated solutions.

\paragraph{Domain Adaptation.} For the backbone model of the reward model, we select {LLaMA-3.1-8B-Instruct}, which is the same choice as for the policy model in our framework.
Although LLMs have demonstrated remarkable performance on various text generation tasks, they still face challenges in complex mathematical reasoning tasks. To address this, we fine-tune the backbone model with mathematical instructions, enabling it to better adapt to mathematical scenarios and tasks. 
Similar to the fine-tuning process for the policy model, we randomly sample 70K problems from the NuminaMath dataset and use Qwen2.5-Math-72B-Instruct to generate solutions for each problem, thereby enhancing the quality of the training data. Incorrect solutions are then removed from the training set, and the remaining high-quality data is used to fine-tune the original model, resulting in a more capable backbone model.

\paragraph{Generative Training.}
The generative RM is trained to evaluate the correctness of the solution based on the given problem, by generating the assessment text in natural language. 
To further leverage the advantages of the generative RM, we design a prompt template that guides the reward model to generate an evaluation of a given solution. The template is as follows:
\begin{promptbox}[Prompt Template for Generative Reward Model.]{codegray}
\label{rm-prompt}
\texttt{\#\#\# Outcome-supervised Reward Model\\Given the problem and the solution from the student, you should verify the correctness of the final answer. It should be noted that the final answer is put in boxed\{\}.\\\\\{Problem\} \{Generated Solution\}\\\\Is the answer correct (Yes/No)?}
\end{promptbox}
For an instance $\langle q_k, \tilde{s}_k, l_k \rangle$ in the training dataset $\mathcal{D}_T$, we utilize the above prompt template to format the question $q_k$ and the generated solution $\tilde{s}_k$, and train the backbone model to learn to generate the assessment $l_k$.
Formally, the objective function of generative reward model training is as follows:
\begin{equation}
\label{eq-rm-train}
    \mathcal{L}_{\text{GRM}}=-\sum_{\langle q_k, \tilde{s}_k, l_k \rangle \in \mathcal{D}_T} P(g(l_k) | q_k, \tilde{s}_k),
\end{equation}
where $g(l_k)$ means that the output is formatted as a text following the above template. 
After training the backbone model on training dataset $\mathcal{D}_T$ through objective function Eq.~\ref{eq-rm-train}, we can obtain the generative reward model $\mathcal{R}$.

\paragraph{Active Learning.} 
To further enhance the ability of the reward model, we employ active learning~\cite{Margatina2023active} in our training process, selecting the high-quality, hard instances to improve the training of the reward model.
First, we follow the training approach introduced above to derive an initial reward model, denoted as $\widetilde{\mathcal{R}}$.
Then, we utilize $\widetilde{\mathcal{R}}$ to score the generated solutions in the original training dataset $\mathcal{D}_{O}$, and rank these generated solutions in descending order based on the reward scores from $\widetilde{\mathcal{R}}$.
\ignore{\begin{equation}
    \text{Score}_{i,j} = \widetilde{\mathcal{R}}(q_i, \tilde{s}_{i,j}).
\end{equation}} 
After scoring, we split the responses for each problem into two sets according to the correctness label (\ie correct solution set and incorrect solution set). We then rank the responses in each set in descending order based on the reward scores and finally select the top-ranked solutions from both sets, respectively. This allows us to effectively select high-score correct and incorrect responses: high-score correct responses are more likely to be of high quality, while high-score incorrect responses represent challenging samples that the original reward model cannot reliably identify. These two types of data are useful for enhancing the quality and difficulty of the training dataset, thereby improving the performance of the reward model. Finally, we follow the data cleaning process introduced above to remove similar solutions and select an equal number of correct and incorrect responses. These selected samples are used to construct the active learning dataset $\mathcal{D}_{A}$, and we train the backbone model on this dataset $\mathcal{D}_{A}$ to obtain the improved reward model, which is denoted as  $\mathcal{R}$.

\paragraph{Reward Normalization.} 
To assign the reward scores, we first obtain the probabilities of the reward model $\mathcal{R}$ generating the tokens of ``\texttt{Yes}'' and ``\texttt{No}'' (over the entire vocabulary), denoted as $p_{Y}$ and $p_{N}$, respectively. 
Then, to ensure $p_{Y}+p_{N}=1$, we normalize the original probability scores  as follows: 
\begin{equation}
    \tilde{p}_{Y}=\frac{e^{p_{Y}}}{e^{p_{Y}}+e^{p_{N}}},~~\tilde{p}_{N}=\frac{e^{p_{N}}}{e^{p_{Y}}+e^{p_{N}}}.
\end{equation}
After normalization, we can see that $\tilde{p}_{Y}\in[0,1]$ and a higher probability can be regarded as the higher probability that the given solution is more likely to be correct, and we use the absolute score of the RM to guide the search process.

\subsubsection{Iterative Training for Mutual Evolution}
\label{sec:iterative_training}
In our framework, the policy model and the reward model are two highly interrelated components: their training involves using data generated or selected by one another, and they jointly perform the reasoning process to arrive at the final solution. Consequently, we explore the possibility of mutual improvement for these two components and iteratively refine the model capacities. 

Specifically, let $\mathcal{P}_0$ and $\mathcal{R}_0$ denote the initial policy model and reward model (after format and domain adaptation), respectively. 
At the $i$-th iteration, policy model $\mathcal{P}_{i-1}$ first generates the candidate solutions based on the problems in the training dataset.
Then, these generated solutions and problems are used to compose the original training dataset for training the reward model $\mathcal{R}_{i-1}$ and obtain an improved reward model $\mathcal{R}_{i}$.
Next, we leverage the new reward model $\mathcal{R}_{i}$ to score the original training dataset and perform preference optimization on policy model $\mathcal{P}_{i-1}$ to obtain the improved policy model $\mathcal{P}_{i}$. 
Finally, the new policy model $\mathcal{P}_{i}$ and reward model $\mathcal{R}_{i}$  will be utilized as the backbone model for the next iteration.
During the training process, $\mathcal{R}_{i}$ is trained on the data generated from $\mathcal{P}_{i-1}$. 
Therefore, the policy model can perform step-by-step reasoning and searching under the effective guidance of the reward model.

\subsection{Search Algorithm}
In this section, we first describe the overall process of search algorithm using the specially trained policy model and reward model~(Section~\ref{sec_search_overall}), followed by two improvements for search performance (Section~\ref{sec_search_opt}). 

\subsubsection{Overall Process}
\label{sec_search_overall}
We perform the search process by executing a certain number of search steps. Each search step consists of four main operations: selection, expansion, simulation, and backpropagation. Next, we introduce each operation in detail.  

\paragraph{Selection.} {We first consider the selection method that follows the standard MCTS algorithm~\cite{cameron2012mcts}: } starting from the root node, the algorithm traverses the tree by selecting child nodes based on the Upper Confidence Bound~(UCB) value, which is a widely used criterion that balances exploration and exploitation. 
Let $s_t$ denote the $t$-th node. 
The UCB value of a node to be selected is determined by two key variables: $V(s_{t+1})$, the reward value of the candidate child node given by the reward model $\mathcal{R}$, and $N(s_t)$, the visit frequency of $s_t$. At each step, the child node with the highest UCB would be chosen. This process can be formally  represented as:
\begin{equation}
    s_{t+1} = \underset{s_{j}\in \text{Children}(s_t)}{{\arg\max}} [V(s_{j})+c \cdot \sqrt{\frac{\log N(s_t)}{1+N(s_{j})}}], 
\end{equation}

where $c$ is a constant that determines the level of exploration. This operation repeats until a leaf node is reached. 
For selection, MCTS locally selects from the child nodes of the current node, while we further design another selection method that considers all the leaf nodes of the current search tree. 
Concretely, the algorithm begins by collecting all leaf nodes of the current search tree and calculating the average value $\mu$ and standard deviation $\delta$ of their reward values. A threshold $p$ is then dynamically calculated based on $\mu$ and $\delta$. Finally, nodes with a reward value $V$ exceeding $p$ will be selected. The calculation method is outlined as follows:
\begin{align}
    \mu &= \frac{1}{N_{\text{leaf}}}\sum_{s \in \text{leaf}} V(s),\\
    \delta &= \sqrt{\frac{1}{N_{\text{leaf}}}\sum_{s \in \text{leaf}} \big( V(s) -\mu\big)^2},\\
    p &= \mu + \lambda\cdot \delta,
\end{align}
where $\lambda$ is a constant that specifies the extent to which $p$ exceeds $\mu$, and $N_{\text{leaf}}$ denotes the number of leaf nodes in the current search tree.
We refer to this modified selection method as {MCTS$_G$}, and will further compare the performance of MCTS and MCTS$_G$ in Section~\ref{subsec-search_algorithm}.

\paragraph{Expansion.} 
\label{sec_search_overall_expansion}
If the selected leaf node is not in a terminal state (\ie  the final answer has not been generated), the node will be expanded in depth with $k$ child nodes $\{ s_c \}$, by decoding for one step using the policy model. The initial values of these newly expanded child nodes will be determined through the following simulation operation.

\paragraph{Simulation.} This operation employs rollout samples to estimate the potential value of the expanded child node $s_{c}$. Concretely,  
from the selected child node $s_{c}$, the policy model $\mathcal{P}$ performs $n$ rollouts~(\ie generates a complete solution $\tau^{(i)}$ based on the current node's state). The reward model $\mathcal{R}$ then evaluates the correctness of each rollout trajectory by computing a scalar value. The evaluation results are then averaged to determine the initial reward value $V$ of the selected child node. This process can be formally represented by the following equation:
\begin{equation}
    V(s_{c})=\frac{1}{n}\sum_{i=1}^n \mathcal{R}(\tau^{(i)}). 
\end{equation}

\paragraph{Backprogation.} Once the initial reward values of the expanded child nodes $\{s_c\}$ are evaluated, they are propagated back up the tree. Each node along the path updates its statistics, such as the visit count $N$ and the state value $V$, which will inform future selections. The process can be formally defined as: 
\begin{align}
V(s_t) &\longleftarrow \frac{N(s_t)\cdot V(s_t)+ \sum^{k} V(s_{c})}{N(s_t)+k},\\
N(s_t) &\longleftarrow N(s_t)+k. 
\end{align}

These operations are repeated for a predetermined number of iterations or until a time limit is reached, allowing the algorithm to build a question-dependent search tree. 
Additionally, to accelerate the search process and leave more exploration space for subsequent search at the beginning of the simulation, we also implement pre-expansion before starting the simulation. 

\paragraph{Pre-expansion.} Pre-expansion involves expanding a significant number of nodes in the initial layers of the search tree before the standard search process begins. Instead of selecting the most promising leaf node and expanding it, we expand all nodes in the initial layers, to enhance the coverage of correct actions. 
After this expansion, the algorithm proceeds with the aforementioned operations of selection, expansion, simulation, and backpropagation.

\paragraph{Discussion on Search Algorithms.}   
The essence of incorporating search algorithms in the reasoning framework is to help effectively and efficiently expand the solution space that LLMs can explore. In principle, we can employ various search algorithms to improve the reasoning performance of LLMs.  
In this work, we mainly consider three search algorithms: MCTS, MCTS$_G$, and beam search. 
MCTS has been widely used in various games~\cite{AlphaGo} and has now been adapted to instruct the reasoning of LLMs. It focuses on analyzing the most promising actions from the current state and expanding the search tree based on random sampling of the search space. However, it may produce suboptimal solutions due to its local selection strategy, and the inherent randomness may affect search effectiveness. 
To potentially overcome this limitation,  MCTS$_G$ expands the selection scope to all the leaf nodes of the current tree. However, 
the algorithm may waste computational resources by pursuing incorrect paths, leading to a deeper exploration of high-value paths in complex or error-prone problems.
In comparison, beam search is a simplified search algorithm that explores a fixed number of the best candidates at each step. While it benefits from parallelizability, which accelerates the search process, it may yield suboptimal results due to its limited exploration.

\subsubsection{Performance Optimization}
\label{sec_search_opt}
In this part, we discuss the optimization strategies for improving the performance of the search framework. 

\paragraph{Self-consistency Enhancement with Rollout Samples.} During the tree search process, we perform rollouts to estimate the value of nodes, which results in a large number of samples with complete solutions. These samples can be leveraged to improve reward evaluation. Based on this idea, in addition to relying on the reward model to score the steps generated by the model, we also calculate the self-consistency~(SC) score of the answers produced by the model itself, a proven effective indicator in existing studies~\cite{xue2023sc}. Specifically, when calculating the reward for each sample $\tau^{(i)}$, we combine the reward model’s score with the SC score {derived from all historical rollout samples}:
\begin{equation}
\mathcal{R}^{+}(\tau^{(i)}) = (1 - \alpha) \cdot \mathcal{R}(\tau^{(i)}) + \alpha \cdot \text{SC}(\tau^{(i)}), 
\end{equation}
where a weighted factor $\alpha$ is applied to the SC score that returns the proportion of the answer of $\tau^{(i)}$ in the historical rollout samples. 
{We empirically find that SC often provides accurate predictions of the ground-truth answer, particularly when the question difficulty is relatively low. Overall, it serves as a valuable indicator for assessing the correctness of candidate answers. } 

\paragraph{Tool Manipulation.}
In each problem-solving step, the policy model may introduce  calculation errors. While these errors may appear insignificant individually, they can accumulate and disrupt the solution process if left unchecked. Unlike logical mistakes, these  miscalculations often go undetected by reward models and self-consistency checks, as they are not specially designed to assess the accuracy of numerical calculations. To address this limitation, we integrate a calculator tool into our framework, ensuring precise, step-by-step verification of calculation results. 
After extracting the equations from each generated step, we implement the calculator tool by using the SymPy library~\cite{aaron2017sympy} to recompute these expressions. This allows us to obtain the correct answers and generate feedback signals that indicate whether the results match those in the generated steps.
By verifying and correcting each equation, the tool prevents basic computational errors, ensuring that the search process remains focused on valid solution paths and logical problem-solving.
\section{Experiments}

In this section, we conduct experiments to examine the effectiveness of the implemented framework.  

\subsection{Evaluation Setup}
To demonstrate the effectiveness of our framework, we conduct experiments on four challenging mathematical benchmarks: MATH-OAI~\cite{LightmanKBEBLLS24lets}, GSM-Hard~\cite{GaoMZ00YCN23pal}, OlympiadBench~\cite{HeLBHTSHHHZLQL024olympiad}, and College Math~\cite{TangZWW24mathscale}. 
The test set sizes for the four benchmarks are 500, 1319, 675, and 2818, respectively. To save testing time, we randomly sample 500 samples from each of the last three benchmarks for evaluation.
We select LLaMA-3.1-8B-Instruct~\cite{llama3} as the backbone model for both the policy and reward models because it demonstrates excellent overall capabilities and does not saturate these benchmarks.
For each benchmark, we employ the same evaluation tools as those used in prior research~\cite{Yang2024qwen2.5math} and report the average performance of various methods over all test problems. 
Aside from the main experiments, we primarily report results on the MATH-OAI dataset, unless otherwise specified.

\subsection{Main Results}

\subsubsection{Overall Performance Comparison}

As the main experiments, we compare four methods based on the same backbone model: zero-shot CoT~(\textbf{CoT}) with fine-tuned policy model, 
best-of-$N$ chosen by the reward model~(\textbf{BoN}, $N=100$), and our tree search based reasoning framework~(\textbf{STILL-1}). Specially, We utilize Llama-3.1-8B-Instruct LLM as the backbone model for policy model and reward model. And we utilize the same prompt for all methods as described in Section~\ref{sec:format_adaptation}.

We first present the overall performance comparison of the four methods on the selected evaluation benchmarks. As shown in Table~\ref{tab:main_res}, the CoT strategy can improve the performance of the original chat model to some extent, as these test tasks require enhanced step-by-step reasoning capabilities.
Among these methods, search-based approaches (\ie BoN and SKILL-1) deliver more superior performance because they effectively expand the solution space for challenging reasoning tasks. Finally, our framework achieves the best performance among all methods, {enhancing the reference baseline by 46.9\%, 7.3\%, 91.6\%, and 31.4\%} on the four evaluation benchmarks, respectively. These results indicate that our framework effectively enhances the reasoning capabilities of LLMs on complex mathematical tasks.

\begin{table}[htbp]
    \centering
    \small
    \setlength\tabcolsep{4.0pt}
    \label{tab:main_res}
    \caption{Performance comparison of different methods on four benchmarks. {``baseline'' refers to the CoT reasoning method based on the original chat model without any further training}, while ``\emph{w/} CoT'' is implemented in a similar way but with specific training as in Section~\ref{sec-policy}. The \textbf{bold} fonts denote the best performance among the compared methods, and we also report the gain over the baseline performance  (in percentage).}
      \begin{tabular}{l | cccccccc}
      \toprule
      \textbf{Method} & \multicolumn{2}{c}{\textbf{MATH-OAI}} & \multicolumn{2}{c}{\textbf{GSM-Hard}} & \multicolumn{2}{c}{\textbf{OlympiadBench}} & \multicolumn{2}{c}{\textbf{College Math}} \\ 
      \cmidrule(r){2-3}\cmidrule(r){4-5}\cmidrule(r){6-7}\cmidrule(r){8-9}
       & \textbf{Acc (\%)} & \textbf{Gain (\%)} & \textbf{Acc (\%)} & \textbf{Gain (\%)} & \textbf{Acc (\%)} & \textbf{Gain (\%)} & \textbf{Acc (\%)} & \textbf{Gain (\%)} \\
        \midrule
      baseline &48.2  & - &  38.4& - & 17.9 & -&  34.1& -\\
       \midrule
      \emph{w/} CoT  & 58.3  & +21.0 &  38.5& +0.3 & 19.2 & +7.3 &  39.0& +14.7 \\
       \emph{w/} BoN  & 69.0  & +43.2 &  38.8& +1.0 & 30.3 & +69.3 & 43.0 & +26.0\\
       \emph{w/} STILL-1 & \textbf{70.8}  & \textbf{+46.9} & \textbf{41.2} & \textbf{+7.3} & \textbf{34.3} & \textbf{+91.6} &  \textbf{44.8} & \textbf{+31.4} \\
      \bottomrule
      \end{tabular}
      \label{tab:main_res}
\end{table}

\subsubsection{Results of Iterative Training}
{In this section, we examine how the performance of the policy model and reward model evolves during the multi-turn iterative training process. As described in Section~\ref{sec:preference_optimization} and Section~\ref{sec:data_construction}, both the training of the policy and reward models involves data selection based on the reward model. Therefore, in this part, we conduct a comparison experiment with random selection to highlight the superiority of the reward-based selection method. For the detailed analysis of the selection strategies of the reward model, we conduct further ablation experiments in Section~\ref{sec:data_selection_rm}. We execute two iterations and present the results in Table~\ref{tab:preference_selection}.}

First, it can be observed that the reward-based selection method outperforms random selection in improving the performance of both the reward and policy models. This indicates that feedback from the reward model not only facilitates its own activate learning (Section~\ref{sec:data_selection_rm}) but also aids in the preference optimization of the policy model (Section~\ref{sec:preference_optimization}). Second, across two iterations, we can observe continual improvements for both policy and reward models on almost all datasets, suggesting that they can mutually enhance each other during the iterative training, which demonstrates the effectiveness of our framework.

\begin{table}[t]
    \caption{The effect of different data strategies for selecting positive and negative preference pairs. The \textbf{bold} fonts denote the best performance. We report the accuracy of \textbf{CoT} and \textbf{BoN} for the policy and reward model, respectively.}
    \setlength\tabcolsep{4.0pt}
    \small
    \centering
    \begin{tabular}{ccccccccc}
        \toprule
        \multirow{3}*{\textbf{Data Selection}} & \multicolumn{4}{c}{\textbf{Random}} & \multicolumn{4}{c}{\textbf{RM Selected}}\\
        \cmidrule(r){2-5}\cmidrule(r){6-9}
         & \textbf{\makecell{MATH\\OAI}} & \textbf{\makecell{GSM\\Hard}} & \textbf{\makecell{Olympiad\\Bench}} & \textbf{\makecell{College \\Math}} & \textbf{\makecell{MATH\\OAI}} & \textbf{\makecell{GSM\\Hard}} & \textbf{\makecell{Olympiad\\Bench}} & \textbf{\makecell{College \\Math}}\\
        \midrule
        \multicolumn{9}{c}{\textit{Policy Model} (CoT)} \\
        Iteration 1 & 54.4 & 37.8 & 19.0 & 36.2 & 58.3  & {38.5} & 19.2 & 39.0 \\
        Iteration 2 & 53.4 & 36.0 & 20.8 & 35.2 & {58.5} & 36.3 & {21.5} & {40.0} \\
        \midrule
        \multicolumn{9}{c}{\textit{Reward Model} (BoN-10)} \\
        Iteration 1 & 62.6 & 37.0 & 27.2 & 41.0 & 64.8 & 37.7 & 28.1 & 41.5 \\
        Iteration 2 & 64.6 & 37.8 &  27.6 & 41.7 & {65.4} &  {38.3} & {28.5} & {41.8}  \\
        \bottomrule
    \end{tabular}
    \label{tab:preference_selection}
\end{table}

\subsection{Further Analysis of Policy Model Training}

In this part, we focus on examining the effect of policy model training. We adopt three evaluation metrics: {\texttt{accuracy} (ratio of the correctly solved test problems by direct generation), \texttt{maj@10} (ratio of the correctly solved test problems by majority vote among ten generated solutions), and \texttt{pass@10} (recall of the correctly solved test problems among ten generated solutions)}. 
The latter two are regarded as coverage metrics for the generated solutions. Next, we give detailed results and analysis.

\begin{table}[htbp]
 \setlength\tabcolsep{3.0pt}
    \centering
    \small
    \caption{The performance of policy model on four benchmarks by using different synthesis model for reading format adaptation. The \textbf{bold} fonts denote the best performance of performing CoT generation with greedy search.}
      \begin{tabular}{l|cccc}
      \toprule
        \textbf{Synthesis Model} &  \textbf{\makecell{MATH-OAI}} & \textbf{\makecell{GSM-Hard}} & \textbf{\makecell{OlympiadBench}} & \textbf{\makecell{College Math}} \\ 
  \midrule
      \multirow{3}[3]{*}{Llama-3.1-8B-Instruct} & 46.5 & 35.1 & 14.4 & 31.1 \\ 
       & ~~~~~~~~~56.1{$_{\texttt{maj@10}}$} & ~~~~~~~~~38.7{$_{\texttt{maj@10}}$} & ~~~~~~~~~23.3{$_{\texttt{maj@10}}$} & ~~~~~~~~~37.0{$_{\texttt{maj@10}}$} \\
       & ~~~~~~~~~~75.6{$_{\texttt{pass@10}}$} & ~~~~~~~~~~54.9{$_{\texttt{pass@10}}$} & ~~~~~~~~~~39.8{$_{\texttt{pass@10}}$} & ~~~~~~~~~~50.1{$_{\texttt{pass@10}}$} \\
    \midrule
      \multirow{3}[3]{*}{Qwen2.5-72B-Instruct} & \textbf{54.8} & 38.0 & \textbf{21.9} & 33.1 \\ 
      & ~~~~~~~~~61.2{$_{\texttt{maj@10}}$} & ~~~~~~~~~41.3{$_{\texttt{maj@10}}$} & ~~~~~~~~~26.0{$_{\texttt{maj@10}}$} & ~~~~~~~~~34.6{$_{\texttt{maj@10}}$} \\
      & ~~~~~~~~~~79.8{$_{\texttt{pass@10}}$} & ~~~~~~~~~~59.3{$_{\texttt{pass@10}}$} & ~~~~~~~~~~41.6{$_{\texttt{pass@10}}$} & ~~~~~~~~~~44.5{$_{\texttt{pass@10}}$} \\
    \midrule
      \multirow{3}[3]{*}{Qwen2.5-Math-72B-Instruct} & 54.7 & \textbf{40.5} & 20.8 & \textbf{34.0} \\ 
      & ~~~~~~~~~62.3{$_{\texttt{maj@10}}$} & ~~~~~~~~~41.6{$_{\texttt{maj@10}}$} & ~~~~~~~~~25.1{$_{\texttt{maj@10}}$} & ~~~~~~~~~35.6{$_{\texttt{maj@10}}$} \\
      & ~~~~~~~~~~79.1{$_{\texttt{pass@10}}$} & ~~~~~~~~~~59.5{$_{\texttt{pass@10}}$} & ~~~~~~~~~~42.4{$_{\texttt{pass@10}}$} & ~~~~~~~~~~45.0{$_{\texttt{pass@10}}$} \\
      \bottomrule
      \end{tabular}
\label{synthesis model}
\end{table}

\subsubsection{Effect of Data Synthesis Model}

Recall that we use a more capable model for data synthesis. We now examine the effect of different data synthesis models by comparing LLaMA-3.1-8B-Instruct, Qwen2.5-72B-Instruct~\cite{qwen2.5}, and Qwen2.5-Math-72B-Instruct~\cite{Yang2024qwen2.5math}, which represent the policy model itself, a strong general-purpose model, and a strong domain-specific model, respectively. We fix the total amount of formatted synthetic data at 10K and compare the performance of using different models. As shown in Table~\ref{synthesis model}, the results indicate that self-generated data offers little improvement and may even lead to performance degradation, whereas data generated by strong models significantly enhances performance. Additionally, data synthesized by the domain-specific model does not show significant differences compared to the general-purpose model, with slight improvements observed on certain datasets.

These results indicate two key points: First, when the foundational model has strong capabilities, a higher-quality solution process is necessary to further enhance its complex reasoning abilities. This can be achieved by distilling higher-quality instruction data using a larger LLM. Second, the help from a larger general LLM is comparable to that of a larger domain-specific LLM for the policy model, reducing reliance on the latter.

\label{sec_policy}
\begin{figure}[htbp]
    \centering
    \includegraphics[width=1\linewidth]{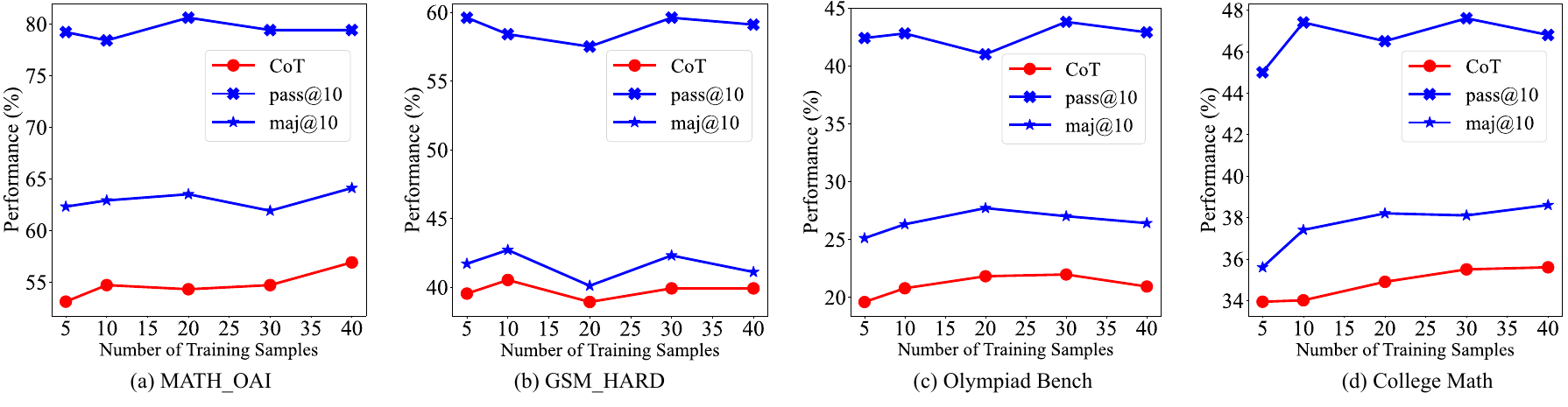} 
    \caption{Performance of policy model on four benchmarks by using different amounts of training data for reasoning format adaptation. ``CoT'' means the CoT generation with greedy search.}
    \label{fig:scaling}
\end{figure}

\subsubsection{Effect of Training Data Scaling}
We further investigate how the amount of synthetic data affects model performance. Specifically, we conduct experiments using varying amounts of synthetic data from the Qwen2.5-Math-72B-Instruct, including 5K, 10K, 20K, 30K, and 40K, while keeping the other settings fixed. Figure~\ref{fig:scaling} displays the tuning results achieved with different amounts of synthetic data. {Compared to the original policy model, using more synthetic data for format adaptation overall enhances the reasoning performance of the policy model}. 
However, as the amount of training data increases, the performance gains become less significant or may even decline. On the other hand, the model's \texttt{pass@k} fluctuates within a narrow range, which does not bring significant improvements. 

Considering the balance between the cost of synthetic data and downstream performance, we utilize 10K training data for reasoning format adaptation in our framework.

\begin{figure}[htbp]
    \centering
    \includegraphics[width=0.6\linewidth]{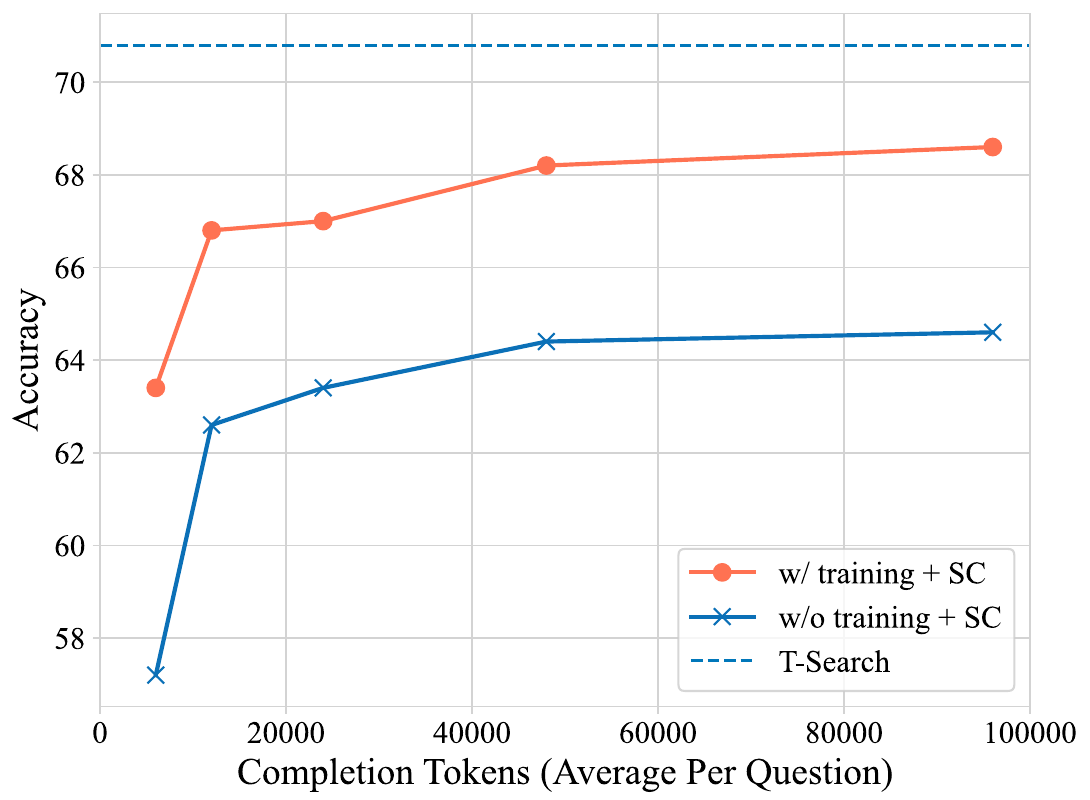} 
    \caption{Self-consistency results  \emph{with} and \emph{without} optimization using our training method.}
    \label{fig:sc_policy}
\end{figure}

\subsubsection{Performance Improvement on Self-consistency}

To further evaluate the effectiveness of our training method for the policy model, we conduct self-consistency experiments to assess how an improved policy model contributes to model performance using the self-consistency~(SC) strategy. We select the SC strategy because we empirically find that it performs very well when employing a sufficient number of rollouts. Specifically, we compare the policy model \emph{with} and \emph{without} (\ie the original model) optimization using our training method, while varying the sampling budget~(\ie the number of rollouts). Additionally, we include the performance of the overall reasoning framework (\ie STILL-1 in Table~\ref{tab:main_res}) as a reference. 
From Figure~\ref{fig:sc_policy}, we can see that our training method significantly enhances SC performance by improving the underlying policy model, highlighting its effectiveness. Moreover, SC with the enhanced policy model acts as a very strong baseline, achieving high performance yet still underperforming compared to our entire reasoning framework.

\subsection{Further Analysis of Reward Model Training}
\label{exp-rm}
In this section, we conduct detailed analysis on the impact of different training strategies for the reward model.

\subsubsection{Effect of Model Adaptation Strategies}
Building on the original backbone model, we can employ different adaptation strategies to enhance its performance as a reward model. Next, we examine two main strategies:  domain adaptation and  format adaptation.

\paragraph{Domain Adaptation.}
As detailed in Section~\ref{reward-model-training-approach}, the backbone model was fine-tuned with mathematical instructions to adapt more effectively to the specific requirements of mathematical problem-solving. The results in Table \ref{tab:initialization_approach} indicate  models fine-tuned with mathematical instructions generally achieve higher performance, as measured by accuracy with  BoN-$N$ (with $N$ best candidates), as $N$ increases. This suggests that incorporating mathematical domain knowledge can enhance the model's ability to assign more precise rewards, thereby improving assessment accuracy.

\paragraph{Format Adaptation.} 
Section~\ref{sec:format_adaptation} introduces the fine-tuning approach for aligning the policy model with the desired step-by-step reasoning format. We can perform similar adaptation for the reward model.
As shown in Table~\ref{tab:initialization_approach}, format adaptation seems to be helpful to improve the original model when $N=5,10$. However, as $N$ increases, the benefits become relatively limited, and it does not lead to consistent improvement as observed with domain adaptation.
\begin{table}[t]
    \caption{The effect of different adaptation strategies.}
    \centering
    \begin{tabular}{lcccccc}
        \toprule
        \textbf{Model} & \textbf{BoN-2} & \textbf{BoN-3} & \textbf{BoN-5} & \textbf{BoN-10} & \textbf{BoN-20} & \textbf{BoN-40}\\
        \midrule
        backbone & \textbf{60.8} & \textbf{63.8} & 65.6 & 64.4 & 64.4 & 63.8 \\
         \emph{w/} domain  adaption & 60.0 & 62.8 & 65.2 & \textbf{64.8} & \textbf{65.6} & \textbf{66.6} \\
        \emph{w/}  format adaption & 59.8 & 63.8 & \textbf{66.4} & \textbf{64.8} & 65.4 & 65.4 \\
        \bottomrule
    \end{tabular}
    \label{tab:initialization_approach}
\end{table}

\subsubsection{Effect of Data Selection Strategies}
\label{sec:data_selection_rm}
In Section~\ref{sec:data_construction}, we design specific strategies to clean and select the data samples for training the reward model. 
Next, we conduct experiments to examine the effectiveness of these strategies.

\begin{table}[t]
    \caption{The effect of different data cleaning strategies.}
    \centering
    \begin{tabular}{cccccccc}
        \toprule
        \textbf{Dedupilicate} & \textbf{Debias} & \textbf{BoN-2} & \textbf{BoN-3} & \textbf{BoN-5} & \textbf{BoN-10} & \textbf{BoN-20} & \textbf{BoN-40} \\
        \midrule
         \checkmark & \checkmark & 60.0 & 62.8 & \textbf{65.2} & \textbf{64.8} & \textbf{65.6} & \textbf{66.6} \\
         \checkmark & \ding{55} & 59.2 & 62.4 & 64.6 & 62.4 & 63.6 & 62.8 \\
         \ding{55} & \checkmark & 60.0 & \textbf{63.2} & \textbf{65.2} & 63.0 & 64.6 & 64.0 \\
        \bottomrule
    \end{tabular}
    \label{tab-data_cleaning_methods}
\end{table}

\paragraph{Data Cleaning.} In the data cleaning process, we first remove similar responses to improve the diversity of the training data (``\emph{Deduplicate}'') and then reduce dataset bias by selecting an equal number of positive and negative responses  (``\emph{Debias}''). We conduct experiments to train the reward model on datasets without removing lexically similar responses or without reducing data distribution bias and present the results in Table~\ref{tab-data_cleaning_methods}. From Table~\ref{tab-data_cleaning_methods}, we can observe that these data strategies can improve the performance of the reward model. Furthermore, comparing the results between Deduplicate and Debias, we find that removing the latter strategy leads to a larger performance drop, as LLMs might overfit to the ``shortcuts'' present in datasets with biased data distribution.

\paragraph{Data Selection.}
\begin{table}[t]
    \caption{The effect of different data selection strategies.}
    \small
    \centering
    \begin{tabular}{cccccccc}
        \toprule
        \multirow{2.5}*{\textbf{Positive}} & \multirow{2.5}*{\textbf{Negative}} & \multicolumn{3}{c}{\textbf{Iteration 1}} & \multicolumn{3}{c}{\textbf{Iteration 2}}\\
        \cmidrule(r){3-5}\cmidrule(r){6-8}
         & & \textbf{BoN-5} & \textbf{BoN-10} & \textbf{BoN-20} & \textbf{BoN-5} & \textbf{BoN-10} & \textbf{BoN-20} \\
        \midrule
        Random & Random &61.2 & 62.6 & 63.2 & 65.2 & 64.8 & 65.6 \\
        Descending ($\downarrow$) & Descending ($\downarrow$) & \textbf{62.2} & \textbf{64.8} & \textbf{66.4} & \textbf{66.2} & \textbf{65.4} & \textbf{67.0} \\
        Ascending ($\uparrow$) & Descending ($\downarrow$) & 58.8 & 61.0 & 61.8 & 64.0 & 60.6 & 60.2 \\
        Descending ($\downarrow$) & Ascending ($\uparrow$) & 61.8 & 63.4 & 64.0 & 65.0 & 64.6 & 65.2 \\
        Ascending ($\uparrow$) & Ascending ($\uparrow$) & 58.6 & 60.4 & 61.0 & 65.6 & 64.0 & 62.8 \\
        \bottomrule
    \end{tabular}
    \label{tab-data_active_learning}
\end{table}
To evaluate the effectiveness of active learning in data selection, we conduct experiments using different sample ranking methods and report the results in Table~\ref{tab-data_active_learning}. Ranking both positive and negative samples in descending order of scores achieved the best performance compared to other ranking methods and demonstrates a significant improvement over random sampling. For positive samples, data selection in ascending order significantly reduces the reward model's effectiveness, highlighting the importance of high-quality samples for learning. In contrast, for negative samples, selecting high-scoring ones is more beneficial, as it introduces greater discrimination difficulty. Thus, choosing samples with higher scores overall helps construct a high-quality dataset.

\subsubsection{Effect of Model Design and Training}
In this part, we examine the impact of different designs and training methods for the reward model.

\paragraph{Objective Function.}
\begin{table}[t]
    \caption{Performance comparison of different objective functions for discriminative reward models.}
    \centering
    \small
    \begin{tabular}{lccc}
        \toprule
        \textbf{Objective Function} & \textbf{BoN-5} & \textbf{BoN-10} & \textbf{BoN-20} \\
        \midrule
        $\mathcal{L}_1=-(\log{(\sigma(y^{+}_i))} + \log{(1-\sigma(y^{-}_i))}$ & \textbf{56.4} & \textbf{57.2} & \textbf{59.8} \\[7pt]
        $\mathcal{L}_2=\sigma(y^{+}_i)-\sigma(y^{-}_i)$ & 41.6 & 41.4 & 41.8 \\[7pt]
        $\mathcal{L}_3=(\sigma(y^{+}_i)-1)^2+(\sigma(y^{-}_i)-0)^2$ & 55.2 & 56.2 & 57.0  \\[7pt]
        $\mathcal{L}_4=-\log(\sigma(y^{+}_i-y^{-}_i))$ & 54.8 & 56.6 & 58.8 \\
        \bottomrule
    \end{tabular}
    \label{tab-dis_objective_function}
\end{table}

Although we ultimately adopt a generative reward model, we also empirically examine the performance of discriminative reward models with different objective functions for optimization. We add a linear head to the policy model to project its hidden state into a numerical score and implement scoring-based objective functions (\ie $\mathcal{L}_1$ and $\mathcal{L}_3$) and ranking-based objection functions (\ie $\mathcal{L}_2$ and $\mathcal{L}_4$) to train the discriminative reward model.
The experimental results are shown in Table~\ref{tab-dis_objective_function}.
According to the evaluation results, the binary cross-entropy (BCE) $\mathcal{L}_1$ outperforms the other functions, since it is well suited for quality assessment tasks with binary-labeled training data. 
Additionally,  $\mathcal{L}_3$ and $\mathcal{L}_4$ achieve comparable performance on the scenarios with fewer candidates (\eg BoN-5, BoN-10), while $\mathcal{L}_4$ perform well when the number of candidates increases (\eg BoN-20).
However, since $\mathcal{L}_4$ is the ranking-based objective function, it cannot provide an absolute score for generated content. Therefore, in scenarios requiring absolute scores (\eg MCTS), $\mathcal{L}_4$ may not be appropriate.

\paragraph{Backbone Model.}
\begin{table}[t]
    \caption{Performance comparison of different backbone models for reward models.}
    \centering
    \small
    \begin{tabular}{lccccccc}
        \toprule
        \multirow{2.5}*{\textbf{Models}} & \multicolumn{3}{c}{\textbf{Discriminative Reward Model}} & \multicolumn{3}{c}{\textbf{Generative Reward Model}} \\
        \cmidrule(r){2-4}\cmidrule(r){5-7}
        & \textbf{BoN-5} & \textbf{BoN-10} & \textbf{BoN-20} & \textbf{BoN-5} & \textbf{BoN-10} & \textbf{BoN-20} \\
        \midrule
        LLaMA 3.1 8B Instruct & 58.6 & 57.2 & 60.8 & 57.6 & 60.0 & 62.4  \\
        LLaMA 3.1 70B Instruct & 61.8 & 63.2 & 64.2 & 63.0 & 65.8 & 67.2  \\
        Qwen2.5 Math 7B Instruct & 63.4 & 67.6 & 70.6 & 63.6 & 67.4 & 71.2 \\
        \bottomrule
    \end{tabular}
    \label{tab-dis_backbone}
\end{table}

The selection of the backbone model is another key factor for reward model training.
In this experiment, we examine three types of LLMs as the backbone model, including small-scale general LLM (\ie LLaMA-3.1-8B-Instruct), large-scale general LLM (\ie LLaMA-3.1-70B-Instruct\cite{llama3}), and the small-scale math-specific LLM (\ie Qwen2.5-Math-7B-Instruct\cite{Yang2024qwen2.5math}).
We conduct the experiment on both discriminative and generative reward models, and present the results in Table~\ref{tab-dis_backbone}.
Overall, generative reward models outperform discriminative reward models, suggesting that the generative paradigm can more effectively leverage the original capabilities of the backbone model, as it maintains a similar training approach. Additionally, increasing the model size can also enhance the assessment capability. Furthermore, the math-specific model outperforms the large-scale model, emphasizing the importance of the backbone model's proficiency in mathematical reasoning.

\subsubsection{Performance of Reward Model on Process-supervised Tasks}
\label{sec:performance_reward_model_diff}
\begin{table}[t]
    \caption{Performance comparison of reward model on different search methods.}
    \centering
    \small
    \begin{tabular}{ccccccc}
        \toprule
        \textbf{CoT} & \multicolumn{3}{c}{\textbf{Depth-First Search}} & \multicolumn{3}{c}{\textbf{Beam Search}}\\
        \cmidrule(r){2-4}\cmidrule(r){5-7}
        - & $N=5$ & $N=10$ & $N=20$ & $N=5,B=5$ & $N=10,B=5$ & $N=20,B=5$ \\
        \midrule
        58.3 & 59.8 & 58.6 & 61.8 & 62.6 & 64.0 & 62.4 \\
        \bottomrule
    \end{tabular}
    \label{tab:process_supervised_performance}
\end{table}

As discussed in Section~\ref{sec-rm_key_issue}, we primarily focus on outcome-supervised reward models due to the intensive computational resources required to generate process-supervised samples. In this part, we demonstrate that the resulting outcome-supervised reward model not only performs well on outcome-level tasks but also possesses some degree of step-level assessment capability.

\paragraph{Evaluation Setting.}    
To evaluate the model's step-level assessment capability, we mainly consider two evaluation tasks based on depth-first search and beam search, respectively. For depth-first search, at each iteration, the model samples $N$ candidate steps based on the given problem and the existing solution steps. The reward model assigns a score to each candidate step, selecting the one with the highest score, which is then appended to the current solution. This process is repeated iteratively, sampling and scoring the subsequent steps until the final solution is reached. For  beam search, the model generates $N$ candidate steps at each iteration, with the reward model selecting the top $B$ steps based on their scores. During the entire search process, only the top $B$ candidates with the highest scores are retained, progressively constructing the final solution. The performance of the reward model on these tasks is evaluated based on the accuracy of the final solutions generated by the model, reflecting its process supervision capability.

\paragraph{Model Performance.}  
We present the results of our reward model on process-supervised tasks in Table~\ref{tab:process_supervised_performance}. As we can see, our generative reward model outperforms the CoT baseline on both evaluation settings.
Moreover, beam search with our reward model can yield larger improvements, since it involves selecting and retaining more samples at each step. 
This indicates that, despite being trained solely on outcome-level data, our reward model is still capable of providing  process-level supervision signals to some extent. 

\subsection{Further Analysis of Search Algorithm}

In this section, we further analyze the effect of different designs for the search algorithm.

\subsubsection{Comparison of Different Search Algorithms}
\label{subsec-search_algorithm}
\begin{figure}[htbp]
    \centering
    \includegraphics[width=0.8\linewidth]{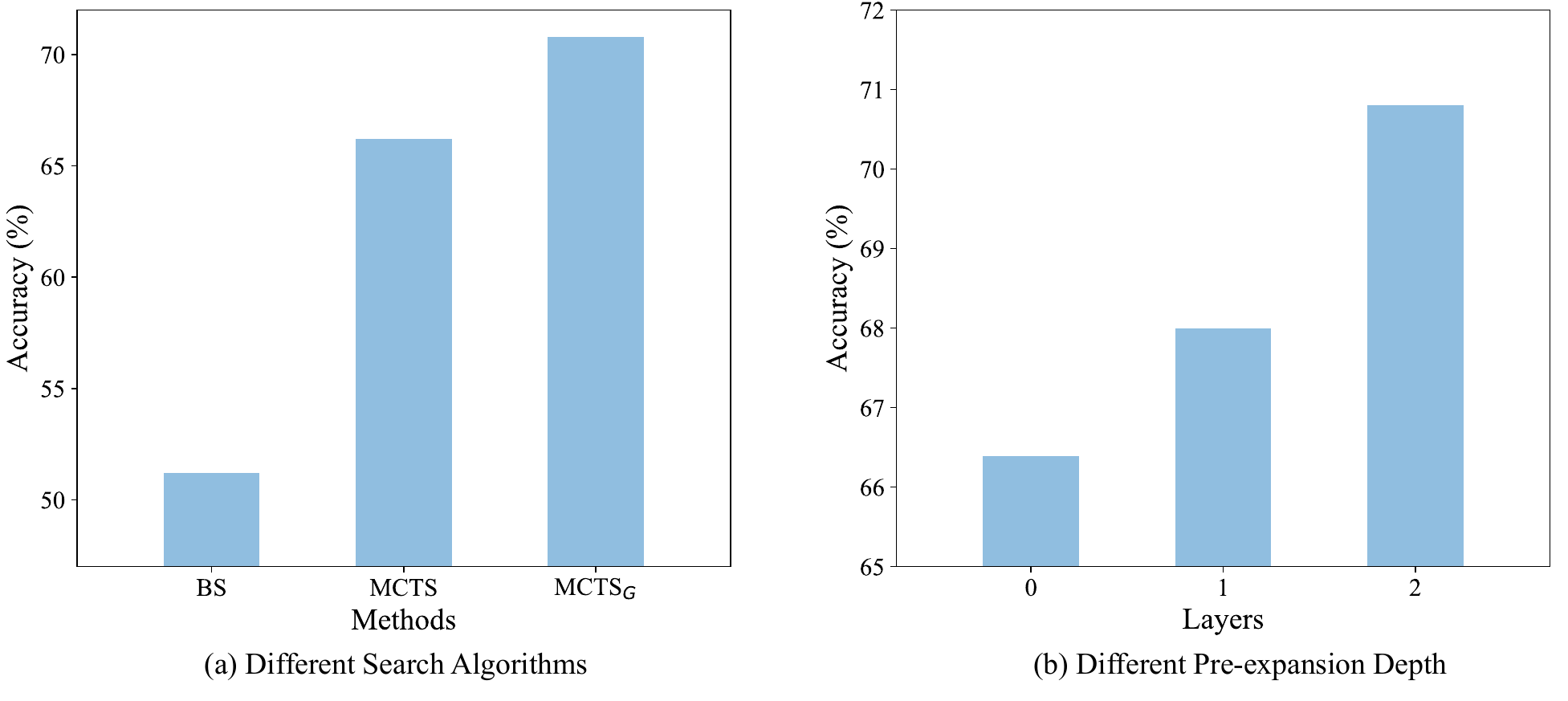} 
    \caption{Effect of search algorithms and pre-expansion depth on model performance.}
    \label{fig:search_res}
\end{figure}

Search algorithms help LLMs explore a broader reasoning space, thereby improving the task performance. In this experiment, we compare the performance of three search algorithms when integrated with our framework, including MCTS, MCTS$_G$, and beam search. 
In MCTS, expansion is performed at a leaf node, which is reached by iteratively selecting the child node with the highest UCB value from the root.
By contrast, in MCTS$_G$, all the leaf nodes are considered for expansion. 
In comparison, beam search employs a simpler strategy by keeping a fixed number of candidates at each step. 
For MCTS and MCTS$_G$, we set the number of expanded child nodes to three; for beam search, we also set the beam size to three, and also conduct rollout to help select top candidates at each step as MCTS. The number of rollouts at each simulation is set to five for the three search algorithms. 
We evaluate each method's accuracy on the MATH-OAI dataset, as shown in 
{Figure~\ref{fig:search_res}~(a)}. 

As we can see, beam search has the lowest accuracy among the three methods. Since beam search only performs forward expansion without the ability to backtrack to previous steps, it can lead to the accumulation of errors that cannot be corrected. 
In contrast, by expanding nodes based on UCB values, MCTS dynamically balances exploration and exploitation, reducing error accumulation and avoiding premature commitment to suboptimal paths, especially for complex problems. However, the extensive exploration can sometimes lead to a slight decrease in accuracy.
Finally, MCTS$_G$, on the other hand, achieves the highest accuracy among the methods tested by considering all the leaf nodes as candidates for selection. While this approach enhances accuracy, it may occasionally lead to deeper exploration to high-value paths in complex or error-prone problems, where the time required for backtracking and path adjustment can exceed that of MCTS.

\ignore{\begin{table}[htb]
    \caption{Performance comparison of different search algorithms on MATH-OAI dataset.}
    \centering
    \small
    \begin{tabular}{lcc}
        \toprule
        \textbf{Methods} & \textbf{Time Cost (s/problem)} & \textbf{Accuracy} \\
        \midrule
        Beam search & 149.02 & 51.20 \\
        MCTS & 237.56 & 66.20\\
        MCTS$_G$ & 329.10 & 70.80 \\
        \bottomrule
    \end{tabular}
    \label{tab:search-method}
\end{table}
}

\subsubsection{Effect of Pre-expansion}

Since the initial steps usually involve understanding the problem and preparing the prerequisites,  the reward model's ability to differentiate reasoning steps at the early stages is  limited. Pre-expansion allows for rapid parallel expansion of the search tree, which is useful to expand the initial  exploration space. In this part, we examine the effect of pre-expansion on model performance. 
{Figure~\ref{fig:search_res}~(b)} presents the performance  by varying the number of pre-expansion layers from zero (no pre-expansion) to two. The results show that pre-expansion can enhance the accuracy compared to the implementation without pre-expansion, underscoring the effectiveness of this technique in optimizing the search process. Considering the computational overhead from pre-expansion, we find that pre-expansion with two layers overall achieves an optimal balance between computational cost and search efficiency.

\ignore{
\begin{table}[htb]
    \caption{Varying the number of pre-expansion layers in our search algorithm.}
    \centering
    \small
    \begin{tabular}{ccc}
        \toprule
        \textbf{Layer} & \textbf{Time Cost (s/problem)} & \textbf{Accuracy} \\
        \midrule
         0 & 388.66& 66.39\\
         1 & 333.34& 68.00\\
         2 & 329.10 & 70.80 \\
        \bottomrule
    \end{tabular}
    \label{tab:pre-expand-layer}
\end{table}
}

\section{Conclusion}

In this paper, we presented a technical report documenting our preliminary exploration of a search-enhanced reasoning framework STILL-1 for LLMs. In implementation, our  framework consisted of three main components:  policy model, reward model, and search algorithm. We carefully discussed various key design considerations and conducted extensive explorations of various implementation details. Preliminary results have demonstrated the effectiveness of the implemented framework on the mathematical reasoning tasks.

This work presents our team's preliminary attempts to reproduce o1-like reasoning systems. We extensively reference various existing techniques from the literature and implement and integrate them into this reasoning system. During our explorations, we find it challenging to configure various design factors, integrate different components, and jointly optimize the entire framework. Actually, we are still searching for the ``real'' (or correct) research pathways to reproduce o1-like systems. 
As we anticipate, the ideal technical approach should be more scalable in both training and inference, simple and elegant in principle, and  capable and general as an enhanced reasoning system. We are aware that our current implementation is far from achieving this, and will continue this project in future work to advance research in this area.

\bibliographystyle{unsrt}
\bibliography{ref.bib}

\end{document}